\documentclass[twocolumn,10pt]{jmlr} 

\usepackage{rotating}
\usepackage{longtable}

\usepackage{booktabs}
\usepackage{siunitx}

\usepackage[switch]{lineno}
\usepackage[noend]{algpseudocode}
\usepackage{algorithm}
\usepackage{algorithmicx}
\usepackage{siunitx} 
\usepackage{multirow}
\usepackage{booktabs}
\usepackage{threeparttable}
\usepackage{adjustbox}
\usepackage{subcaption}
\usepackage{mathrsfs}
\usepackage{stfloats}
\usepackage[skip=2pt,font=small,labelfont=bf]{caption}
\usepackage{xcolor}
\usepackage{color}

\usepackage{dpfloat}
\usepackage{pgffor}
\usepackage{etoolbox}
\ifcsdef{subfigure}{}{\newcommand{\subfigure}{}}

\newcommand{\equal}[1]{{\hypersetup{linkcolor=black}\thanks{#1}}}

\newcommand*{\samplesB}{
relative_errors_plots_diabetes_vcu_MAR.png,
relative_errors_plots_diabetes_vcu_MCAR.png,
relative_errors_plots_diabetes_vcu_MNAR.png
}

\newcommand*{\samplesC}{
relative_errors_plots_diabetic_retinopathy_final_MAR.png,
relative_errors_plots_diabetic_retinopathy_final_MCAR.png,
relative_errors_plots_diabetic_retinopathy_final_MNAR.png
}

\newcommand*{\samplesD}{
relative_errors_plots_eeg_eye_state_MAR.png,
relative_errors_plots_eeg_eye_state_MCAR.png,
relative_errors_plots_eeg_eye_state_MNAR.png
}
\newcommand*{\samplesE}{
relative_errors_plots_myocardial_infarction_MAR.png,
relative_errors_plots_myocardial_infarction_MCAR.png,
relative_errors_plots_myocardial_infarction_MNAR.png
}
\newcommand*{\samplesF}{
relative_errors_plots_wisconsin_bc_diagnosis_MAR.png,
relative_errors_plots_wisconsin_bc_diagnosis_MCAR.png,
relative_errors_plots_wisconsin_bc_diagnosis_MNAR.png
}
\newcommand*{\samplesG}{
relative_errors_plots_wisconsin_bc_prognosis_MAR.png,
relative_errors_plots_wisconsin_bc_prognosis_MCAR.png,
relative_errors_plots_wisconsin_bc_prognosis_MNAR.png
}
\theorembodyfont{\upshape}
\theoremheaderfont{\scshape}
\theorempostheader{:}
\theoremsep{\newline}

\SetKwInput{kwPrec}{Procedure}
\SetKwInput{kwRe}{Return}
\title[M-DEW24]{M-DEW: Extending Dynamic Ensemble Weighting to Handle Missing Values}

\author{%
\Name{Adam Catto}\equal{These authors contributed equally}\Email{adam.catto@mssm.edu}\\
\addr Icahn School of Medicine at Mount Sinai, New York, NY, USA
\AND
\Name{Nan Jia}\footnotemark[1]\Email{njia@gradcenter.cuny.edu}\\
\addr CUNY Graduate Center, New York, NY, USA
\AND
\Name{Ansaf Salleb-Aouissi} \Email{ansafsalleb@columbia.edu}\\
\addr Columbia University, New York, NY, USA
\AND
\Name{Anita Raja} \Email{anita.raja@hunter.cuny.edu}\\
\addr CUNY Hunter College, New York, NY, USA
}

\begin{document}

\maketitle
\begin{abstract}

Missing value imputation is a crucial preprocessing step for many machine learning problems. However, it is often considered as a separate subtask from downstream applications such as classification, regression, or clustering, and thus is not optimized together with them. 
We hypothesize that treating the imputation model and downstream task model together and optimizing over full pipelines will yield better results than treating them separately. Our work describes a novel AutoML technique for making downstream predictions with missing data that automatically handles preprocessing, model weighting, and selection during inference time, with minimal compute overhead. Specifically we develop M-DEW, a Dynamic missingness-aware Ensemble Weighting (DEW) approach, that constructs a set of two-stage imputation-prediction pipelines, trains each component separately, and dynamically calculates a set of pipeline weights for each sample during inference time.  We thus extend previous work on dynamic ensemble weighting to handle missing data at the level of full imputation-prediction pipelines, improving performance and calibration on downstream machine learning tasks over standard model averaging techniques. M-DEW is shown to outperform the state-of-the-art in that it produces statistically significant reductions in model perplexity in 17 out of 18 experiments, while improving average precision in 13 out of 18 experiments.
\end{abstract}

\paragraph*{Data and Code Availability}
This paper uses six public health datasets: the EEG Eye State \citep{roesler2013eeg}, the Diabetic Retinopathy \citep{antal2014diabetic}, the Wisconsin Breast Cancer Diagnosis and Prognosis \citep{street1995inductive}, Diabetes 130-US hospitals for years 1999- 2008 \citep{clore2014diabetes}, and the Myocardial Infarction Complications \citep{golovenkin2020myocardial} datasets from the UCI Machine Learning Repository. The code for the approach paper is available at \url{https://github.com/adamcatto/dynamime}
\paragraph*{Institutional Review Board (IRB)}
This research does not involve human subjects, and thus does not require IRB approval.
\section{Introduction}
\label{sec:intro}

Missing value imputation is a crucial preprocessing step for many machine learning problems, notably in supervised machine learning. Consider biomedical and clinical tasks, such as diagnosis, disease progression modeling, and risk stratification. The datasets on which these machine learning models are trained tend to be sparse: many procedures, tests, and measurements are conducted on a small number of samples across the entire dataset. Therefore, successful prediction modeling in these regimes requires methods for handling missing values that are optimal for downstream prediction tasks. While our approach is motivated by clinical informatics problems and we evaluate our approach on healthcare datasets, our approach addresses the missing data problem that is ubiquitous across a plethora of application domains. We describe the road map leading to our approach below.

It is common knowledge that missing values exist in the health, signal processing, and image recognition fields. Hardware malfunction, machines shutdown and human mistakes are inevitable despite investing much effort to avoid them~\citep{emmanuel2021survey}. One strategy for handling those data is to simply delete portions of the data that are missing. However, deleting such data can bias the data in non-MCAR regimes (see section \ref{sec:missing_data}). Furthermore, in real life, a global complete-case analysis is impossible, people may opt instead to delete features with large amounts of missing values.  The issue with this approach is two-fold: (1) it is plausible that most features in the dataset are sparse, so only the most commonplace measurements will be preserved, and (2) perhaps those sparse features are particularly important for certain sub-groups of subjects for whom the data are available, leading to worse predictions. 

Another approach to missing data is imputation, either by a central tendency statistic on the feature (e.g., mean, median), or as a function of the features which are present in that sample (examples include k-nearest neighbors imputation \citep{zhang2012nearest}, multiple imputation via chained equations \citep{white2011multiple}, and random forest-based prediction of feature value \citep{stekhoven2012missforest}). However, imputation of features which are not strongly-correlated with other features has serious potential to misrepresent the samples and bias the feature’s distribution \citep{white2010bias}. The same issues are present with imputation based on central tendency (e.g., imputation to the population mean lowers standard deviation). 

It is not known a priori what methods are best suited for imputing missing values for a given dataset, especially since certain features may be better predicted by one class of models and covariates vs. others. For instance, one feature may be strongly predicted as an additive model of certain other features, whereas another may be strongly predicted by a nearest-neighbor or tree-like algorithm.

{\it We hypothesize that applying a dynamic ensemble weighting (DEW) approach to the full imputation-classification pipeline will lead to reduced classification errors over the standard soft voting / uniform model averaging (UMA) approach.} 
Specifically, we develop and evaluate M-DEW, a novel approach that learns a dynamic ensemble weighting function for a pool of imputation-classifier pipelines on a per-sample basis, taking into account missing values. M-DEW is designed to address the knowledge gap in the current state-of-the-art by mitigating the effects of sample misrepresentation by estimating the suitability of each missing-data-handling technique for a given sample.
Our approach is designed to offload the problem of selecting imputation and downstream prediction modules by simply supplying a set of imputation-prediction pipelines, and learning an optimal weighting of the pipelines' predictions dynamically for each sample.
Our novel contributions are two-fold: (i) an ensembling technique at the level of joint imputation-prediction pipelines, rather than choosing imputation and prediction strategies separately; and (ii) a method for dynamically assigning pipeline contribution weights for each sample during the inference phase, which outperforms simple model prediction averaging. Our approach augments a meta-layer on top of existing pipelines for prediction modeling with missing data, enabling better techniques to ensemble prior approaches. We show that this comes at a relatively small cost on top of uniform averaging, discussed more in Section~\ref{sec:approach}.

We evaluate the capability of our model weighting scheme to correctly order its class probability estimates using standard metrics such as AUROC, but also note that another benefit of learning a model weighting function is to better calibrate the class predictions per-sample, i.e. to reduce the absolute error for each sample. Whereas AUROC measures a pipeline's ability to separate classes, we can measure the capability of a model weighting function (e.g. M-DEW) relative to a baseline (e.g. soft voting) on a sample-by-sample basis by looking at the distribution of per-sample error reductions between weighting schemes. We refer to the sample-wise errors as \textit{model perplexity}.



This paper is organized as follows: Section 2 describes the related work, Section 3 explains our proposed missingness-aware dynamic ensemble weighting (M-DEW) approach, Section 4 presents experimental set up and statistical evaluation of M-DEW compared to uniform model averaging (UMA), and Section 5 is the discussion of conclusions and future work.
\section{Related Works}
\label{sec:related}
Our method develops an ensembling technique to weigh predictions based on missing data patterns. In this section we review the types of missing data that exist, how to handle them with imputation, and existing ensemble learning methods on which we base our novel contributions.

\subsection{Missing Data}
\label{sec:missing_data}
Missing values are categorized into three bins: (i) missing at random (MAR), (ii) missing completely at random (MCAR), and (iii) missing not at random (MNAR) defined in \cite{mack2018types}. MCAR indicates that missingness of a feature does not depend on the values of other features. For instance, one-off glitches in one sensor isolated from other sensors yield MCAR missingness. MAR indicates that missingness of a feature depends on the values of observed covariates. MAR missing values are generated by some (either deterministic or nondeterministic) function, which could be parameterized by e.g., a quantile function (patients below a certain age would not be screened for various age-related cancers) or logistic regression. 
MNAR indicates that missingness of a feature can depend on both the values of other variables as well as whether other variables are missing. This may be rule-based, for instance: on a survey, if questions are asked in multiple languages, then if one question in a given language is left blank, the others will likely be as well. 

\subsection{Data Imputation}

A straightforward method of imputing missing values is to use column-wise central-tendency statistics such as the mean. In this case, the mean of the non-missing values of each column in the dataset is calculated, and any missing value in a column $c$ is imputed to the mean of $c$. However, this approach can severely perturb the distribution of the data, for instance by decreasing variance and artificially concentrating the probability mass function. Furthermore, if a feature can be modeled as a function of other variables, this dependence is washed away and the conditional / joint distributions become corrupted, in addition to an increase in estimation error as compared to, say, parametric estimation based on the values of other features.

Due to this issue with imputation using any central tendency statistic, we may want to approach the problem as a prediction problem using other features. Some options for imputing continuous features are regression and k-nearest neighbors: one can train a (singular, multivariate) regression model with the column to impute as the target column, and the samples which have the feature present as the training set; the model can be used to predict the column values for those samples which are missing it. For categorical variables, one may consider a similarly simple classifier, such as logistic regression. One particularly useful model for imputing missing tabular data is the MissForest algorithm proposed by \cite{stekhoven2012missforest}, which first imputes all missing values in a column to the mean of that column, then for each column with missing values fits a random forest model on the observed portion and predicts the missing values, updating them with the predictions; this process is done over a chosen number of iterations or until some convergence criterion is met. MissForest can simultaneously impute continuous and categorical variables. We note that the idea behind the MissForest approach can be extended to other regression and classification backbones, viz. gradient boosting, and MLPs. However, not all features are most appropriately modeled with a random forest; some may be better modeled by linear models, nearest-neighbor methods, etc. Therefore we may prefer to choose estimators other than random forest, or even an ensemble of other kinds of estimators. There have been attempts to use Generative Adversarial Networks (GANS) and Autoencoders~\citep{psychogyios2022comparison, lee2021contextual} to address missing value imputation. While these methods work well, both of them focused on one specific dataset and use a single imputation technique as opposed to our ensemble approach which adapts the imputer for different sub-populations of the data.




\subsection{Ensemble Methods}
Ensembling \citep{dietterich2002ensemble} is a class of machine learning techniques that aims to combine multiple predictors into a stronger predictor. For example, decision tree-based ensemble techniques such as random forest \citep{breiman2001random}, boosting \citep{schapire1999brief}, and bagging \citep{breiman1996bagging} aim to combat the typically high variance associated with single decision trees, by aggregating weaker classifiers trained on smaller sample sets and/or feature spaces. Ensemble methods can also combine multiple strong predictors into yet a stronger predictor; for instance, voting classifiers \citep{ruta2005classifier} take predictions from multiple (strong) predictors and use majority rule to obtain a final prediction. The stacked generalization algorithm \citep{wolpert1992stacked} trains a meta-estimator on the set of already-trained prediction models, learning an optimal strategy (e.g. neural network weights or decision tree splits) for combining individual model predictions into a final; this meta-estimator can be any estimator that takes a vector as input, such as a logistic regression model or a decision tree. Extreme Gradient Boost (XGBoost)~\citep{chen2016xgboost} is a popular implementation of gradient tree boosting, which is designed to be space- and time-efficient.

\subsection{Dynamic Ensemble Weighting}
Our method builds on previous work in the area of dynamic ensembling~\citep{zhang2019distance,cruz2020deslib}. As a method that is generally applicable to any classification or regression scenario, DEW has been used for time-series forecasting \citep{du2022bayesian} \citep{chowell2020real}, \citep{lu2022weighted}, and learning of non-stationary time-series data representations \citep{yin2015de2}. In opposition to static ensemble techniques such as random forest, gradient boosting, voting, or stacked generalization, dynamic ensemble techniques choose an ensemble of estimators and a weighting scheme on a sample-by-sample basis. The typical flow of a dynamic ensembling algorithm starts with a pool of estimators and a training set provided as input, upon which the competence of each estimator is assessed per sample; in the inference phase, either a subset of estimators or a weighting scheme is chosen based on the expected competence of each estimator. When only the estimator with the highest expected competence is chosen, this is called \textit{dynamic estimator selection} (e.g. dynamic classifier selection); when a subset of estimators is chosen, this is called \textit{dynamic ensemble selection}; when a weighting scheme is adapted for each sample in the inference phase, this is called \textit{dynamic ensemble weighting}. Most similar to our work is \citep{conroy2016dynamic} which is a variant of AdaBoost \citep{freund1999short} \citep{schapire2013explaining} that is designed to handle missing values; the main difference between \cite{conroy2016dynamic} and our approach is that their approach optimizes a single pipeline for all the data while ours is a meta-ensembling approach, which can take the model proposed in \cite{conroy2016dynamic} as one pipeline input among others. \cite{cruz2020deslib} provide a more comprehensive overview of dynamic ensembling methods. However, these dynamic ensembling methods do not take into account the properties of missing data. Our work fills this gap by adapting DEW to accept missing values in the inputs. Whereas these other algorithms operate on complete datasets (i.e. datasets without missing values), our algorithm can handle incomplete datasets and utilize missingness patterns to make better predictions.


\section{Approach}
\label{sec:approach}




\begin{figure*}[!ht]
    \centering
    \includegraphics[width=0.9\textwidth]{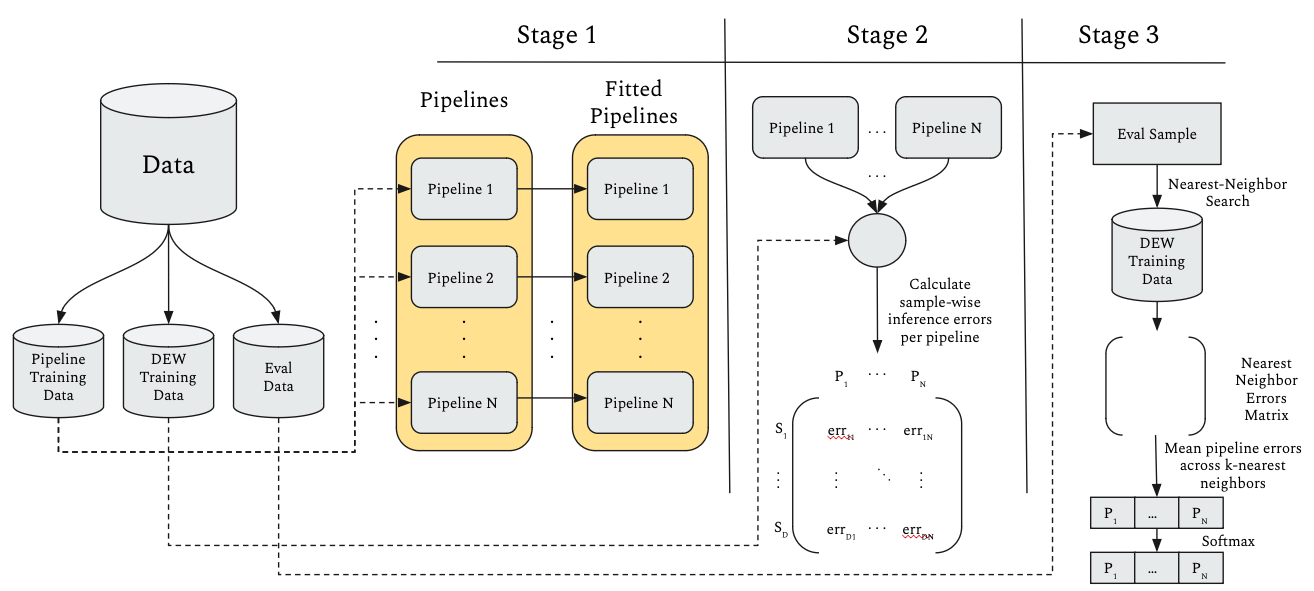}
    \caption{ 
    M-DEW Work Flow Diagram: Phase 1: Imputation and prediction models are fitted to  ``stage-1'' training set. Phase 2:  Inference with each imputation-prediction pipeline run on a ``stage-2'' training set, with  pipeline errors  stored for each sample. Phase 3: Inference on new samples involves weighting  each pipeline's prediction according to its relative competence in the neighborhood of the input sample, i.e. the softmax over pipelines' mean inverse errors.}
    \label{fig:workflow}
\end{figure*}

The Missingness-aware Dynamic Ensemble Weighting (M-DEW) method we have developed is based on the idea that different methods for handling missing data may work better for certain sub-populations of samples than others. 
For instance, within some regions of the input space, certain features may be strongly predicted using a nearest-neighbor approach, but in other regions of the input space certain features may be nonlinear functions of subsets of features, perhaps parameterized by a neural network with Rectified Linear Unit (ReLU) activation or a decision tree. 
{\it The key idea underlying  M-DEW is to dynamically assign weights to prediction models trained using different missing-data-handling techniques, based on the data's missingness patterns, in order to more robustly aggregate predictions from these different models.}

    
  
\begin{algorithm}[ht]
  \caption{Fit Dynamic Ensemble Weighting Model}
  \label{alg:dew_fit}
  \kwPrec{FitM-DEWModel(trainData, trainTargets, valData, valTargets, estimatorPool)}

  sampleWiseEstimErrors $\gets$ dict();
  
  \For{estim in estimatorPool}
      {estim.fit(trainData, trainTargets);
      
      valPredictions $\gets$ estim.estimate(valData); \# list of predictions, one for each sample
      
      sampleWiseEstimErrors[estim] $\gets$ $|$valTargets $-$ valPredictions$|$; \# list of errors, one for each sample}
\kwRe{sampleWiseEstimErrors}
\end{algorithm}

Consider an estimator $\mathscr{E}: \mathbb{F}^n\rightarrow \mathscr{T}^m$ mapping $n-$tuples over a field $\mathbb{F}$ to a target space $\mathscr{T}^m$. 
Typically, the domain of the input data we receive is $\mathbb{R}_{\mathscr{N}}^n $, where $\mathbb{R}_{\mathscr{N}}^n = \mathbb{R} \cup \{NaN\}$, $NaN$ is the symbol for ``missing value''. Mechanisms for handling missing data may be thought of as procedures of the form $\mathscr{M}: \mathbb{R}_{\mathscr{N}}^n \rightarrow \mathbb{R}^m$ where $m\leq n$. Such procedures include deletion mechanisms 
– which remove either samples, features, or both which contain missing values – and imputation mechanisms, which populate the $NaN$ entries with some estimates of what those values might look like, either by central tendency measures or some prediction via other features. As such, they prepare the data to be fed to the estimators for training or inference, such that the new data is compatible with the estimators' domains (i.e., no missing values). In this subsection, we will refer to an estimator as a composite procedure $\mathscr{E} \circ \mathscr{M}$, which first maps from $\mathbb{R}_{\mathscr{N}}^n \rightarrow \mathbb{R}^m$ and then from $\mathbb{R}^m \rightarrow \{0, 1\}$. Within the scope of this paper, we will strictly use imputers $\mathbb{I}$ as opposed to the more general class of $\mathscr{M}$ procedures, thus when we use the term $\mathscr{E}$ for ``estimator'', we are in fact referring to a composite procedure $\mathscr{E} \circ \mathbb{I}: \mathbb{R}_{\mathscr{N}}^n \rightarrow \mathbb{R}^m \rightarrow \{ 0, 1 \}$.

    \begin{algorithm}[ht]
    \caption{Dynamic Ensemble Weighting Model Prediction}
    \label{alg:dew_predict}
    \kwPrec{M-DEWModelPredict(sample, trainData, estimatorPool, sampleWiseEstimErrors, nNeighbors)}   
    errors $\gets$ dict();\\
    normalizedCompetences $\gets$ list();\\
    estimPredictions $\gets$ list();\\
    \For{estim in estimatorPool}
        {neighbors $\gets$ missingnessAwareKNNSearch(sample, trainData, nNeighbors);\\
        errors[estim] $\gets$ err for err in sampleWiseEstimErrors when  sample associated with err is in neighbors;\\
          normalizedCompetence$\leftarrow1-$mean(errors[estim]);\\
          normalizedCompetences.append\\(normalizedCompetence);\\
           estimPredictions.append(estim.estimate(sample));\\
        }
        
    weights $\gets$ softmax(normalizedCompetences);
        
    prediction $\gets$ dotProduct(weights, estimPredictions);
        
    \kwRe{prediction}
        
    \end{algorithm}


To fit the model, we first fit a pool of estimators $\mathscr{E}^* = \{ \mathscr{E}_1, \cdots , \mathscr{E}_n \}$ which can handle missing data on a training dataset. Then we compile predictions and corresponding errors for each sample in the training set from each estimator. This process is demonstrated in Algorithm~\ref{alg:dew_fit}. This classification error matrix is used in the next phase for determining which pipelines are most competent in different areas of the input space. 

In the inference phase (demonstrated in Algorithm~\ref{alg:dew_predict}), for a given sample we impute the sample with each pipeline and perform a k-nearest-neighbor search over the stage-2 imputed training set (see Figure~\ref{fig:workflow}), and calculate a normalized error rate for each estimator's predictions in the sample's neighborhood. Each estimator is assigned a competence score for the sample's neighborhood as one minus the normalized error; the competence scores reflect each pipeline's ability to classify the sample with least perplexity.
The competence scores are compiled into a vector and the weights per estimator are calculated as the softmax of the vector. It is well-known that ensemble approaches are interpretable~\citep{liang2022interpretable}. The competence scores in extending M-DEW can be traced back directly to the pipelines’ performances on each of the selected nearest-neighbor samples

 \begin{table*}[htp!]
  \centering 
   \begin{adjustbox}{max width=\textwidth}
   \begin{tabular}{ll*{6}{c}>{\centering\arraybackslash}p{1cm}}
   \toprule
    \textbf{Dataset} & \textbf{\# of instances} & \textbf{\# of features} & \textbf{types of features} & \textbf{data type} &\textbf{is missing data}\\
   \midrule
    EEG Eye State & 14980 & 15 & Integer, Real & Multivariate & False\\ 
    Diabetic Retinopathy & 1151 & 20 & Integer, Real & Multivariate & False \\ 
    Breast Cancer Diagnosis & 569 & 32 & Real & Multivariate & False\\
    Breast Cancer Prognosis & 198 & 34 & Real & Multivariate & True\\
    Diabetes VCU & 100000 & 55 & Integer & Multivariate, Sequential, Time series & True \\
    Myocardial Infarction & 1700 & 124 & Real & Multivariate & True\\
    \bottomrule
  \end{tabular}
  \end{adjustbox}
  \caption{Description of \num{6} datasets}
  \label{tab:dataset setup} 
\end{table*}
\paragraph{Computational Complexity}
We highlight that our approach comes with minimal overhead in time- and space-complexity. Let $n$ be the sample size of the whole dataset and $p$ be the number of pipelines. The only additional memory requirement is a cached matrix of each baseline classifier's error per sample, which has space complexity $O\left( np + nd\right)$; in exact terms, given a dataset split into stage-1 training, stage-2 training, and test sets, for a proportional $s_1, s_2, s_3$ percentage split, the exact space complexity is $s_2 \cdot n\cdot p $. In our experiments, $s_1=0.16$, thus we can approximate the additional space complexity on top of UMA as $\frac{n\cdot p}{6}$. Thus the space complexity is multilinear in the sample size and number of pipelines.

In terms of time complexity, our algorithm operates in two stages: (i) fitting the classification error matrix and (ii) running inference. To fit the classification error matrix requires a running time of $O(np)$ inferences. At inference time on the test set, all that is required is (i) a $k$ nearest neighbor search on the stage-2 training set, (ii) prediction on those $k$ neighbors by $p$ pipelines, and (iii) a mean-reduce operation followed by softmax and scaling of predictions by the calculated weights. For input samples of dimension $d$, the KNN component has time complexity $O(nd + nk)$, the per-sample inference component has time complexity $O(kp)$, the mean-reduce has time complexity $O(pk)$, and the softmax $\rightarrow$ weight scaling composite is $O(p)$. As the number of test-set samples is linear in the dataset size, the time complexity to run the whole pipeline from training to evaluation after fitting the baseline pipelines is

$$O(np) + O(n) * \left[ O(n(d+k) + O(pk) + O(p) \right]$$
$$=O(np) + O(n^2(d+k) + npk)$$
$$=O(n(p + n(d+k)) + pk)$$
$$=O\left(n^2\left[ d+k + \frac{pk}{n} \right]\right)$$

\medskip

The added time complexity is quadratic in the sample size and multilinear in the number of pipelines, nearest-neighbors, and dataset dimension. This presents a tractable cost on top of the standard soft voting. Notably, increasing the size of the ensemble adds only a linear overhead to the whole pipeline, so one can easily scale up the number of pipelines.

One limitation with our approach is nearest-neighbor methods on input spaces that contain missing values themselves; in the current M-DEW algorithm, the nearest-neighbor lookup for each pipeline is done on that pipeline's imputed dataset. In this way the M-DEW algorithm does not directly utilize the pattern of missingness intrinsic to a given sample, opting instead to use imputations of that sample and training samples. Utilization of missingness patterns can be explored by constructing distance functions on spaces which allow missing values.
\section{Experiments}
\label{sec:experiments}

\begin{table*}[htp!]
  \centering 
  \sisetup{round-mode=places, round-precision=3}
   \begin{adjustbox}{max width=\textwidth}
    \begin{tabular}{lrrrrrr}
     \toprule
     & UMA  & M-DEW  & UMA  & M-DEW  & \% Positive  & \% Samples: \\
     &  AP &  AP &  AUROC & AUROC &  Class &  M-DEW Error $<$  \\
     &    &    &   &   &    &   UMA Error \\
    \midrule
    EEG Eye State MCAR & 81.281 & \textbf{81.720} & 83.208 & \textbf{83.573} & 45 & \textbf{71.729} \\
    EEG Eye State MAR & 88.152 & \textbf{88.676} & 89.328 & \textbf{89.817} & 45 & \textbf{81.429} \\
    EEG Eye State MNAR & 82.007 & \textbf{82.389} & 83.763 & \textbf{84.110} & 45 & \textbf{72.804} \\
    \hline
    Myocardial Infarction MCAR & \textbf{84.528} & 84.300 & \textbf{83.769} & 83.550 & 50 & \textbf{62.731} \\
    Myocardial Infarction MNAR & \textbf{89.564} & 89.510 & \textbf{88.667} & 88.640 & 50 & \textbf{72.140} \\
    Myocardial Infarction MAR & \textbf{91.223} & 91.154 & \textbf{90.321} & 90.301 & 50 & \textbf{76.015} \\
    \hline
    Diabetic Retinopathy MNAR & 77.370 & \textbf{77.409} & 72.279 & \textbf{72.414} & 53 & \textbf{58.471} \\
    Diabetic Retinopathy MAR & 79.192 & \textbf{79.200} & \textbf{74.285} & 74.240 & 53 & \textbf{56.560} \\
    Diabetic Retinopathy MCAR & \textbf{77.349} & 77.346 & 72.650 & \textbf{72.657} & 53 & \textbf{57.168} \\
    \hline
    Breast Cancer Diagnosis MCAR & \textbf{98.401} & 98.398 & 98.593 & \textbf{98.594} & 37 & \textbf{77.329} \\
    Breast Cancer Diagnosis MAR & 98.600 & \textbf{98.606} & 98.935 & \textbf{98.940} & 37 & \textbf{82.601} \\
    Breast Cancer Diagnosis MNAR & 98.524 & \textbf{98.546} & 98.864 & \textbf{98.878} & 37 & \textbf{77.329} \\
    \hline
    Breast Cancer Prognosis MCAR & 30.759 & \textbf{30.852} & \textbf{57.700} & 57.602 & 24 & \textbf{62.626} \\
    Breast Cancer Prognosis MNAR & 36.617 & \textbf{38.053} & \textbf{61.758} & 61.618 & 24 & \textbf{62.626} \\
    Breast Cancer Prognosis MAR & 29.792 & \textbf{29.953} & \textbf{53.487} & 53.445 & 24 & \textbf{54.545} \\
    \hline
    Diabetes VCU MCAR & 63.324 & \textbf{63.343} & 67.594 & \textbf{67.608} & 46 & \textbf{56.125} \\
    Diabetes VCU MNAR & 66.485 & \textbf{66.551} & 70.746 & \textbf{70.781} & 46 & \textbf{59.066} \\
    Diabetes VCU MAR & 66.850 & \textbf{66.883} & 71.151 & \textbf{71.169} & 46 & \textbf{56.185} \\
    \bottomrule
    \end{tabular}
    \end{adjustbox}
\caption{Comparison of M-DEW algorithm to Uniform Model Averaging (UMA) algorithm over six datasets using classification metrics:  Average Precision (AP) (Col. 1- 2),  Area Under the Receiver-Operator Characteristic Curve (AUROC) (Col. 3-4),  \% positive class which is percent of samples in positive class i.e. imbalance in dataset (Col. 5) and \% of test-set samples on which M-DEW outperformed uniform model averaging UMA (Col. 6).}
    \label{tab:metrics_0}
\end{table*}
\subsection{Experimental Setup}

We evaluate M-DEW on six health-related datasets (described in Table~\ref{tab:dataset setup}):
~\cite{roesler2013eeg,street1995inductive, antal2014diabetic,clore2014diabetes, golovenkin2020myocardial, street1995inductive}. Five of these are datasets with $< 100,000$ samples and one dataset with $\geq 100,000$ samples. The breast cancer resource \citep{street1995inductive} has two datasets: diagnosis and prognosis. All datasets used in this study are openly available on the UCI Machine Learning Repository \citep{asuncion2007uci}. The criteria for dataset selection were (i) biomedical-related; (ii) contained numerical features (for regression modeling simplicity). A total of \num{18} experiments were conducted, with \num{6} datasets and \num{3} types of missingness. We first create baseline estimators for building the ensemble model for prediction using \num{4} regression methods as imputers and \num{2} classifiers for prediction. Next, we construct a corresponding error matrix to capture the competence score and then finally construct the M-DEW ensemble model. More details about imputers and classifiers can be found in the supplementary documents.


\subsection{Imputers, Classifiers and Missingness} 
We used four types of regression models for imputation:
a k-nearest neighbor (KNN) imputer, a Bayesian ridge regression imputer, an XGBoost regressor, and a random forest regressor. KNN and linear regression models are the two most straightforward and commonly-used learning algorithms in imputation and have popular implementations in scikit-learn \citep{kramer2016scikit}. We chose random forest due to the popularity of the MissForest algorithm and included XGBoost as another strong ensembling approach. For all experiments, the XGBoost models had a max tree depth of 4 and 50 boosting rounds. Each random forest model was similarly built with 50 trees with a max-depth of 4. Downstream of the imputers, we used two types of classifiers to make predictions: XGBoost and random forest. Like the imputers, the XGBoost models had 50 boosting rounds and a max tree depth of 4, and each random forest model was built from 50 trees with a max tree depth of 4. The imputers were implemented using scikit-learn's IterativeImputer and KNNImputer classes. The Bayesian Ridge regression parameters were set to the scikit-learn default. Details of the implementation of all of our models---imputers, classifiers, and M-DEW models---are available in Appendix~\ref{apd:technical appendix}.

To determine M-DEW's performance on different forms of missing data, we randomly introduced synthetic missing values into the datasets to emphasize the various missingness characteristics discussed in Section \ref{sec:related}. This was done for two reasons: (i) not all datasets we selected contain missing values as shown in Table~\ref{tab:dataset setup}, and (ii) to evaluate our model in similar controlled settings, i.e., settings with similar amounts of missingness. In our experiments, for MCAR-type missingness, 30\% of values in the dataset were masked at random. For MAR and MNAR, 30\% of the columns contained missing values at a rate of 30\% of samples, which depended on $\frac{3}{7}$ of the remaining columns via a logistic regression model with randomly assigned weights; in the case of MNAR, the input features to the logistic regression model were masked at random at a rate of 30\%, so that the missing values in the columns to be masked depended also on the missing values in the inputs that generated the mask – a feature not present in MAR, which distinguishes MAR from MNAR. A Python port of the R-misstastic library \citep{mayer2021rmisstastic} was used to systematically introduce missingness to the data. Details of the implementation of all of our models – imputers, classifiers, and DEW models – are available in the appendix.
\begin{figure}[!htbp]
 \floatconts
    {fig:auroc ranking}
    {\caption {Violin Plot of AUROC for 8 standard imputer-estimators, baseline UMA pipeline and M-DEW pipeline's performance rankings. All metrics in M-DEW ranked at the first place with narrower range. The wider the shape got in the plots, the more samples filled in}}
    {\includegraphics[width=1.0\linewidth]{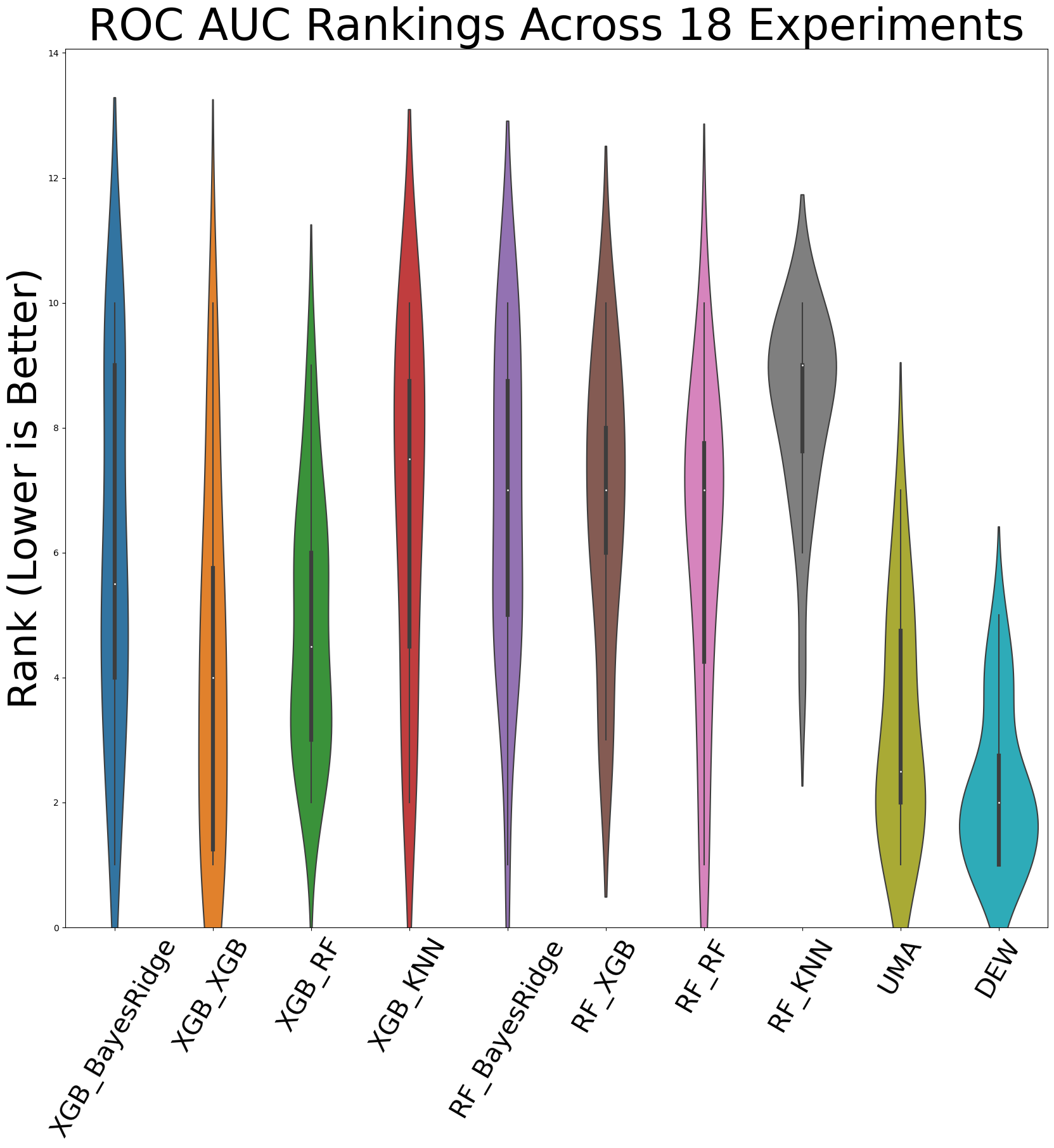}}
\end{figure}

\subsection{Evaluation}  

We compare M-DEW's performance to that of the naive (uniform averaging) ensembling approach using standard classification metrics across six binary classification datasets and three types of missingness (MCAR, MAR, MNAR). 
We report the average precision score and AUROC score across 18 experiments for M-DEW and UMA in Table~\ref{tab:metrics_0}.
Given that a major goal of the M-DEW approach is to reduce the predicted class probability error (i.e. perplexity) compared to the uniform averaging case, we also report the fraction of samples for which M-DEW errors are less than UMA errors, as a measure of the amount of time that M-DEW-calculated weights are directionally correct relative to the baseline.  

In addition to reporting the fraction of samples for which M-DEW improved on UMA, to quantify the statistical significance of these sample-wise error improvements, we run a paired t-test over all samples in each dataset and report the p-value.

\begin{figure}[!htbp]
 \floatconts
    {fig:auroc frequency ranking}
    {\caption {Frequency of rankings that M-DEW has most 1st place ranking among other models}}
    {\includegraphics[width=1.0\linewidth]{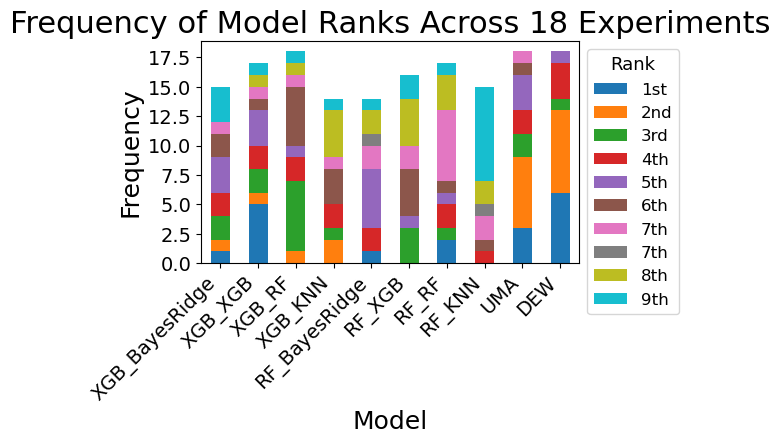}}
\end{figure}
\subsubsection{Main Results}


As shown in Table~\ref{tab:metrics_0}, M-DEW outperforms UMA on the average precision score in 13/18 trials and on AUROC in 11/18 trials. We noted that in cases with more significant class imbalance (breast cancer diagnosis and prognosis), average precision may be a better metric than AUROC, because AUROC can still be quite high while reporting large numbers of false positives; in these cases, M-DEW outperforms UMA on average precision in 5/6 cases, and essentially performs the same in the sixth case. In every experiment, M-DEW classification probability error was globally lower than UMA classification probability error.

In some cases, one of the baseline pipelines may outperform the ensemble methods if it performs more strongly than the other baselines. However, we hypothesized that M-DEW would be robustly among the top-performing models when compared with UMA and the baseline models. To test this, we plotted the relative rankings among the 8 baselines, UMA, and M-DEW on the AUROC metric across the 18 experiments as a violin plot in Figure~\ref{fig:auroc ranking}. Indeed, we see that M-DEW is more consistently better-ranked than the other methods, including UMA; M-DEW's ranking is concentrated heavily among the top 2-3. The other pipelines (including UMA) have median rankings higher than M-DEW, and much higher ranking variance, indicating the stability of M-DEW.

To test M-DEW's ability to improve sample-wise calibration, we run a paired t-test for each dataset that compares the class probability errors between M-DEW and UMA. This significance test confirmed the alternative hypothesis that the M-DEW error distribution has a lower mean than the UMA error distribution's mean. \num{17} out of \num{18} tests indicate that the DEW has a lower error per-sample. 

\begin{table}[htpb]
    \centering
    \footnotesize 
    \begin{adjustbox}{max width=\textwidth}
    \begin{tabular}{|l|l|}
    \hline
        Dataset & P-Value \\
        \hline
        EEG Eye State MCAR & \textbf{$<$ 0.0001}\\
        EEG Eye State MAR & \textbf{$<$ 0.0001}\\
        EEG Eye State MNAR & \textbf{$<$ 0.0001}\\
        \hline
        Myocardial Infarction MCAR & $\mathbf{1.3\times 10^{-07}}$ \\
        Myocardial Infarction MNAR & $\mathbf{1.5\times 10^{-26}}$ \\
        Myocardial Infarction MAR & $\mathbf{1.20\times 10^{-31}}$ \\
        \hline
        Diabetic Retinopathy MNAR & $\mathbf{5.0\times 10^{-05}}$ \\
        Diabetic Retinopathy MAR & \textbf{0.0041} \\
        Diabetic Retinopathy MCAR & \textbf{0.0017} \\
        \hline
        Breast Cancer Diagnosis MCAR & \textbf{0.0002} \\
        Breast Cancer Diagnosis MAR & \textbf{$<$ 0.0001}\\
        Breast Cancer Diagnosis MNAR & \textbf{$<$ 0.0001}\\
        \hline
        Breast Cancer Prognosis MCAR & \textbf{0.0271} \\
        Breast Cancer Prognosis MNAR & \textbf{0.0009} \\
        Breast Cancer Prognosis MAR & 0.1491 \\
        \hline
        Diabetes VCU MCAR & \textbf{$<$ 0.0001}\\
        Diabetes VCU MNAR & \textbf{$<$ 0.0001}\\
        Diabetes VCU MAR & \textbf{$<$ 0.0001}\\
        \hline
    \end{tabular}
    \end{adjustbox}
    \caption{Results of paired t-test;  $p\leq 0.05$ We tested the hypothesis the M-DEW error distribution has a lower mean than the UMA error distribution.}
    \label{tab:paired t-test}
\end{table}

As listed in Table~\ref{tab:paired t-test}, per-sample M-DEW errors are significantly less than the corresponding UMA errors for most samples, in all but one trial. 
Plots of the histograms of relative error improvements can be seen in Figure~\ref{fig:rel_error_hist_eeg-eye-state} and \ref{fig:rel_error_hist_myocardial_infarction}. 
The plots show M-DEW Error minus UMA Error. Therefore, the more dense the plot is to the left of $0$, the better M-DEW performs relative to UMA. We can see that especially for the myocardial infarction and EEG eye state experiments, the distributions are visibly concentrated to the left of zero, which is reflected in the results of paired t-test. For the other datasets, the error difference is concentrated closer to zero, yet still shifted left, indicating that while the magnitude of improvement is lower, M-DEW still tends to be directionally correct. Indeed, in every experiment, M-DEW errors are less than UMA errors for the majority of samples. More details and a thorough discussion of calibration analysis and the Brier score can be found in Appendix~\ref{apd:technical appendix}.
\begin{figure*}[hpt!]
  \floatconts
    {fig:rel_error_hist_eeg-eye-state}
    {\caption{Relative Error Histograms-EEG Eye State}}
    {
      \subfigure[eeg-eye-mar]{\includegraphics[width=0.3\textwidth]{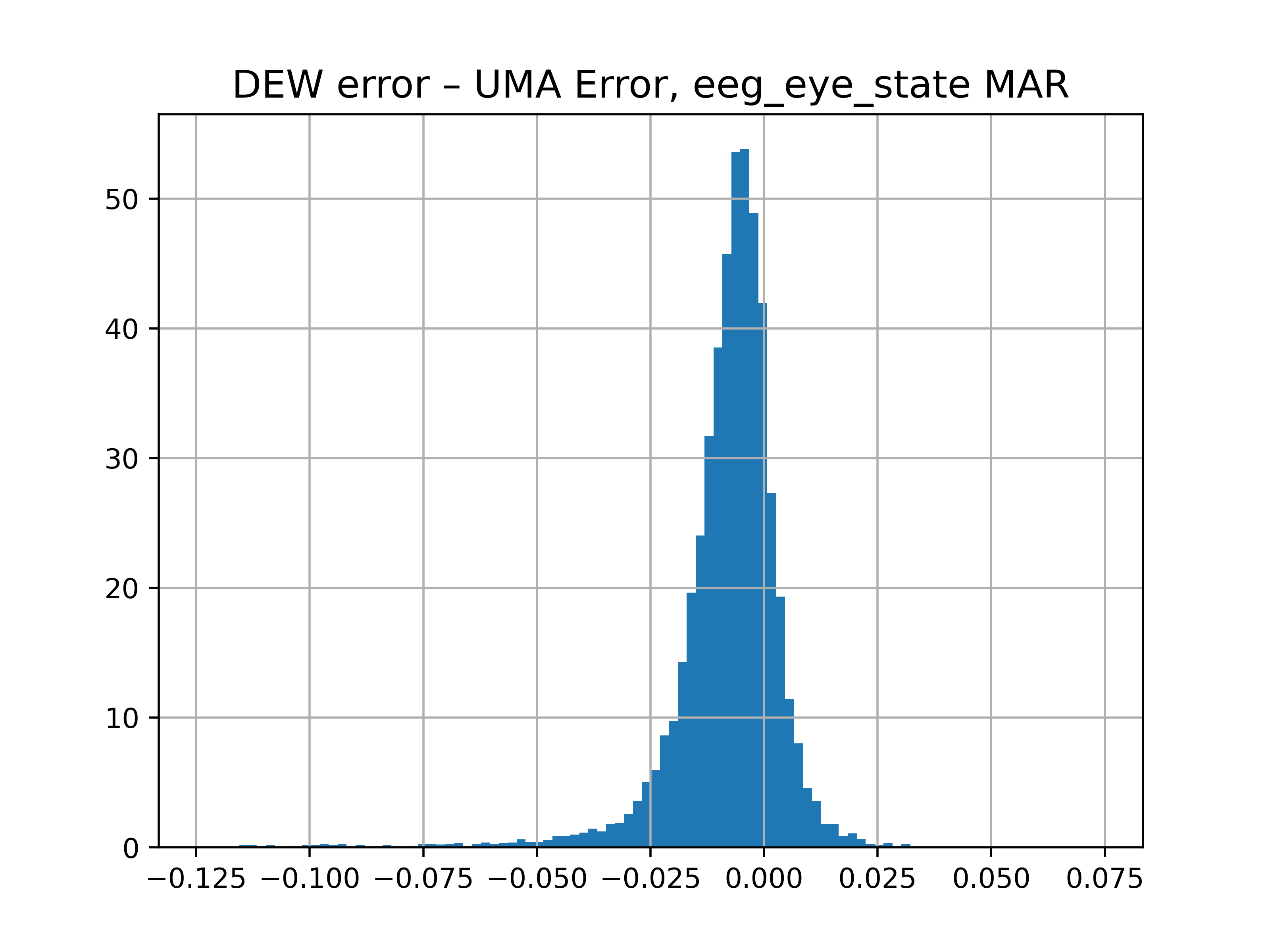}}
      \qquad
      \subfigure[eeg-eye-mcar]{\includegraphics[width=0.3\textwidth]{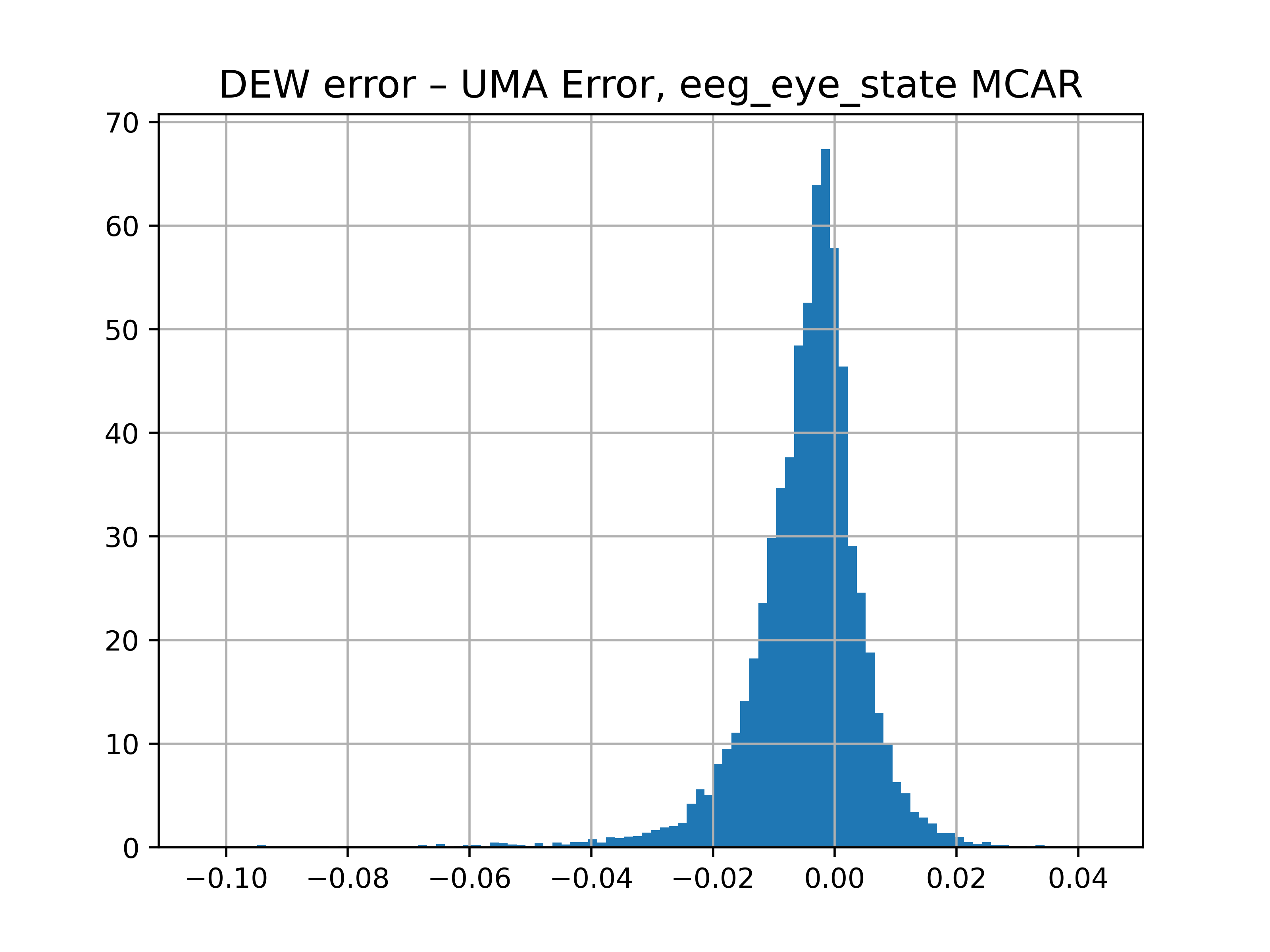}}
      \qquad
      \subfigure[eeg-eye-mnar]{\includegraphics[width=0.3\textwidth]{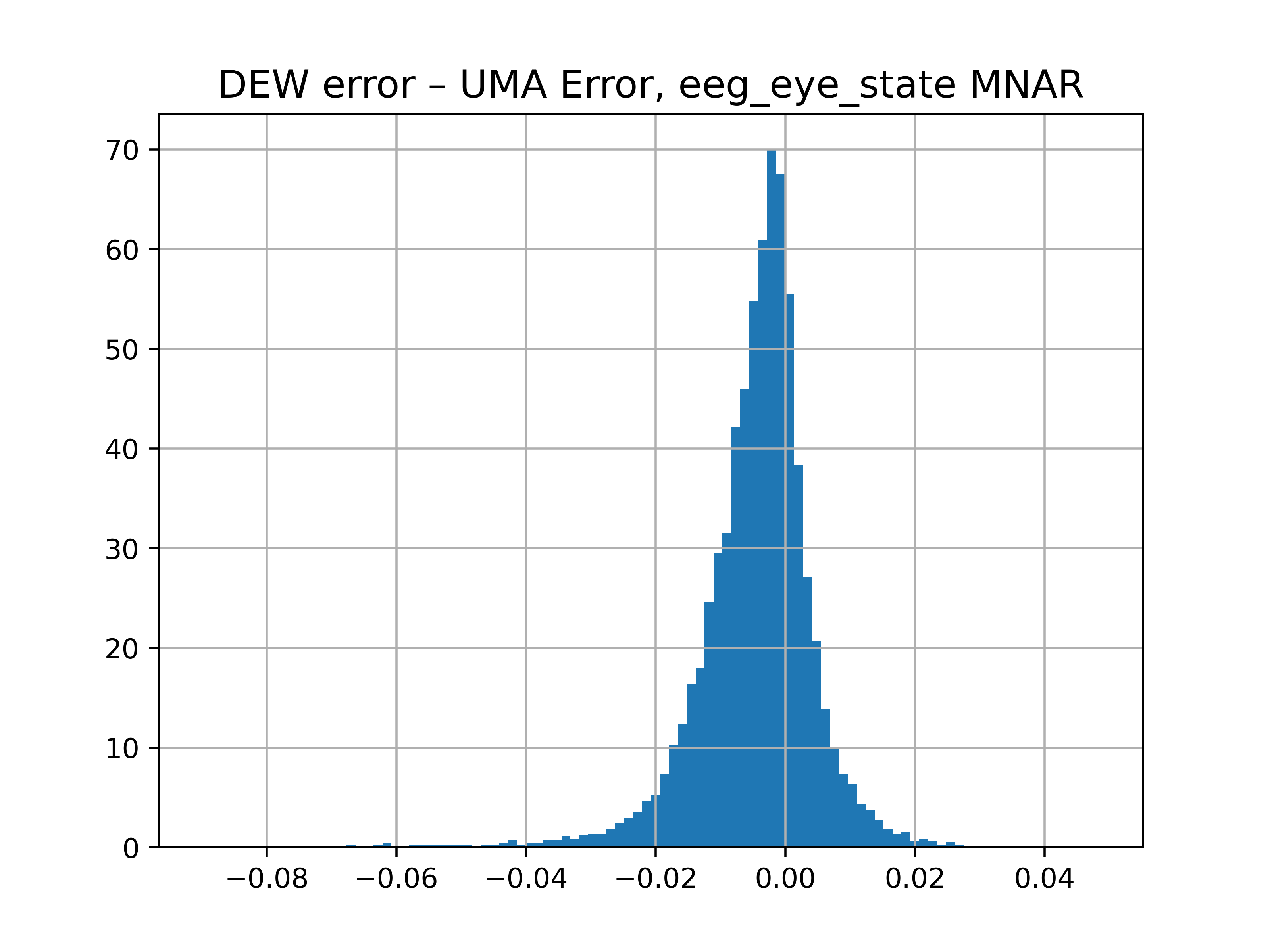}}
    }
\end{figure*}

\begin{figure*}[hptb!]
  \floatconts
    {fig:rel_error_hist_myocardial_infarction}
    {\caption{Relative Error Histograms- Myocardial Infarction}}
    {
      \subfigure[myocardial-infarction-mar]{\includegraphics[width=0.3\textwidth]{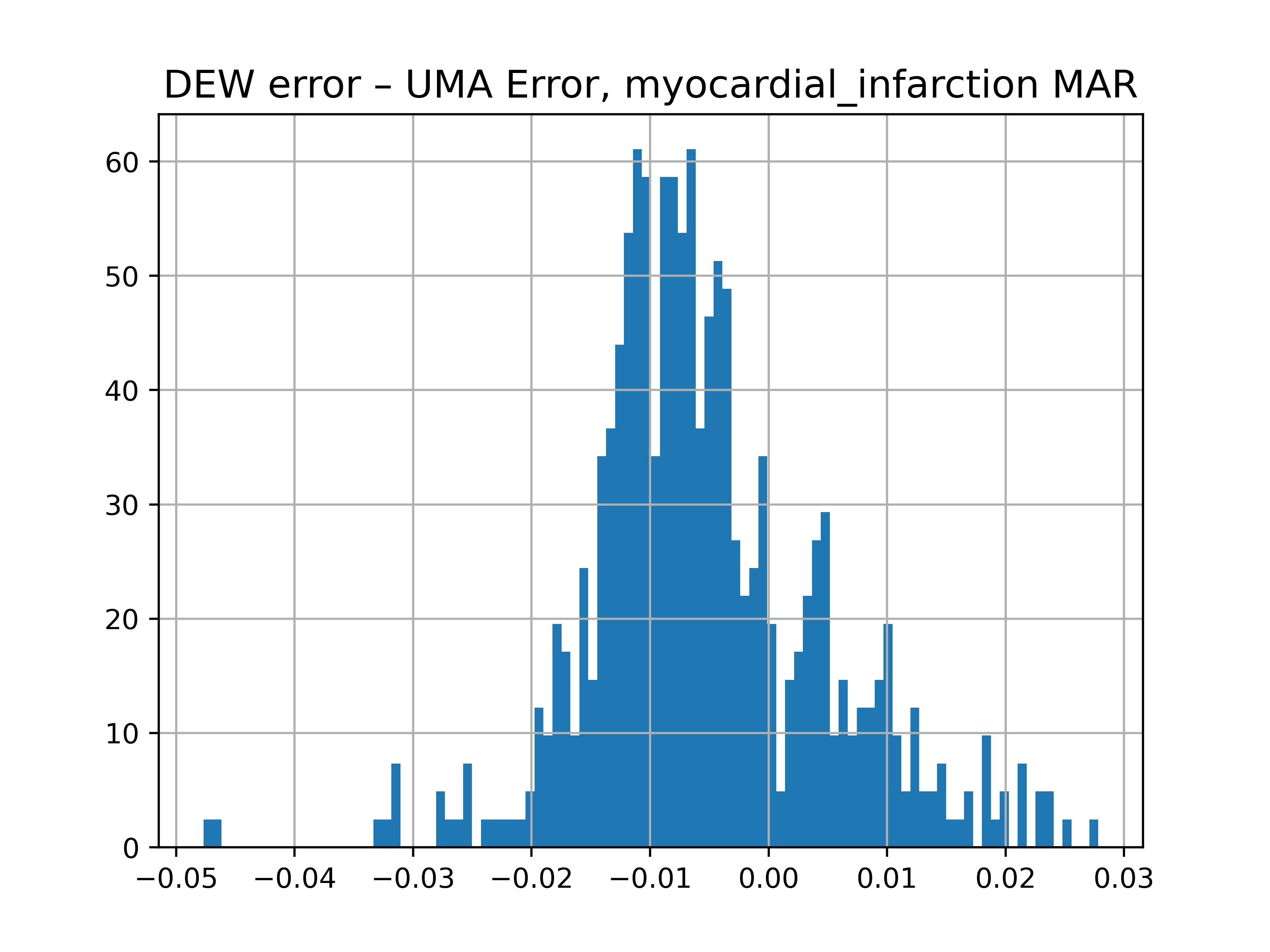}}
      \qquad
      \subfigure[myocardial-infarction-mcar]{\includegraphics[width=0.3\textwidth]{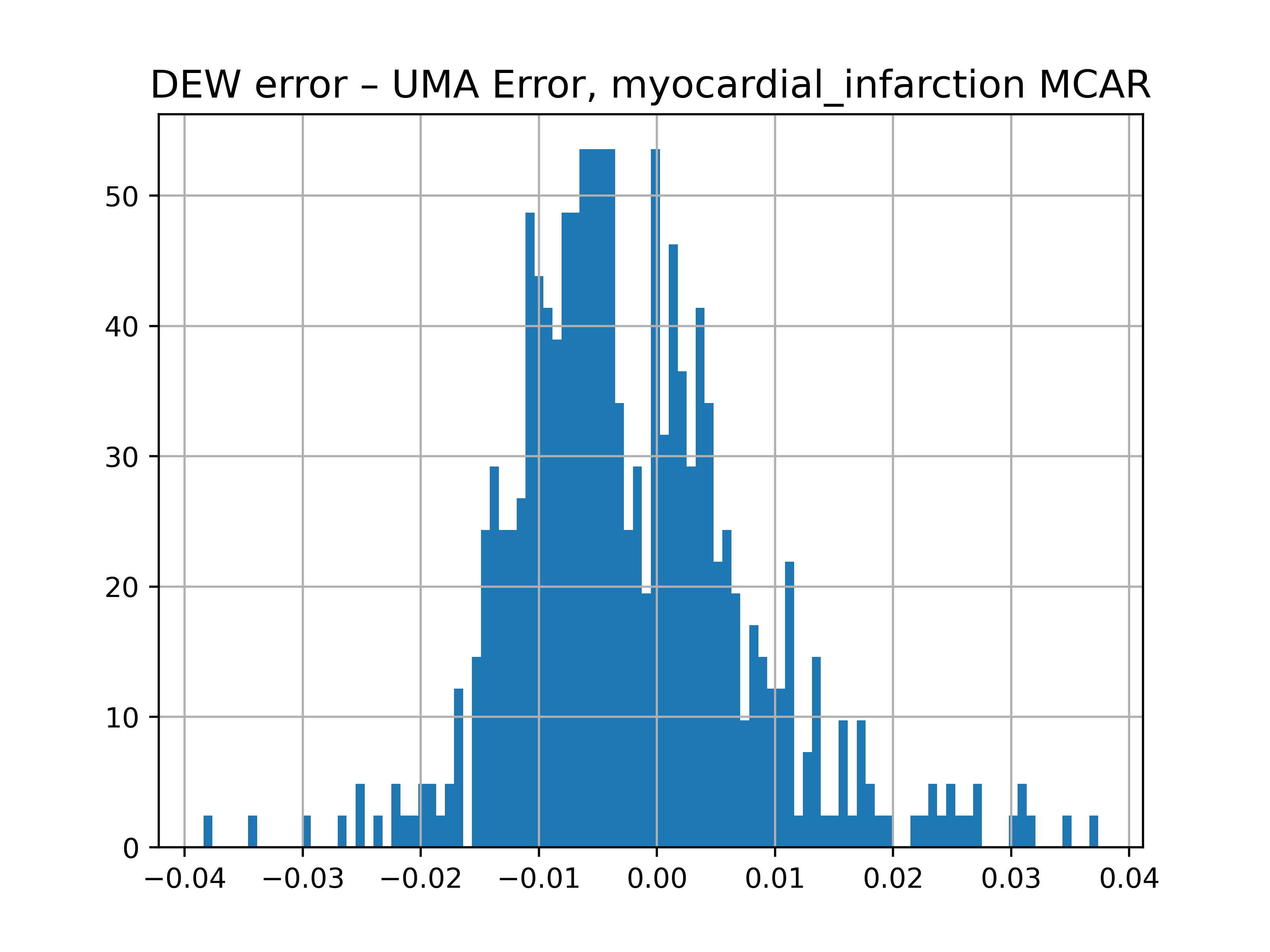}}
      \qquad
      \subfigure[myocardial-infarction-mnar]{\includegraphics[width=0.3\textwidth]{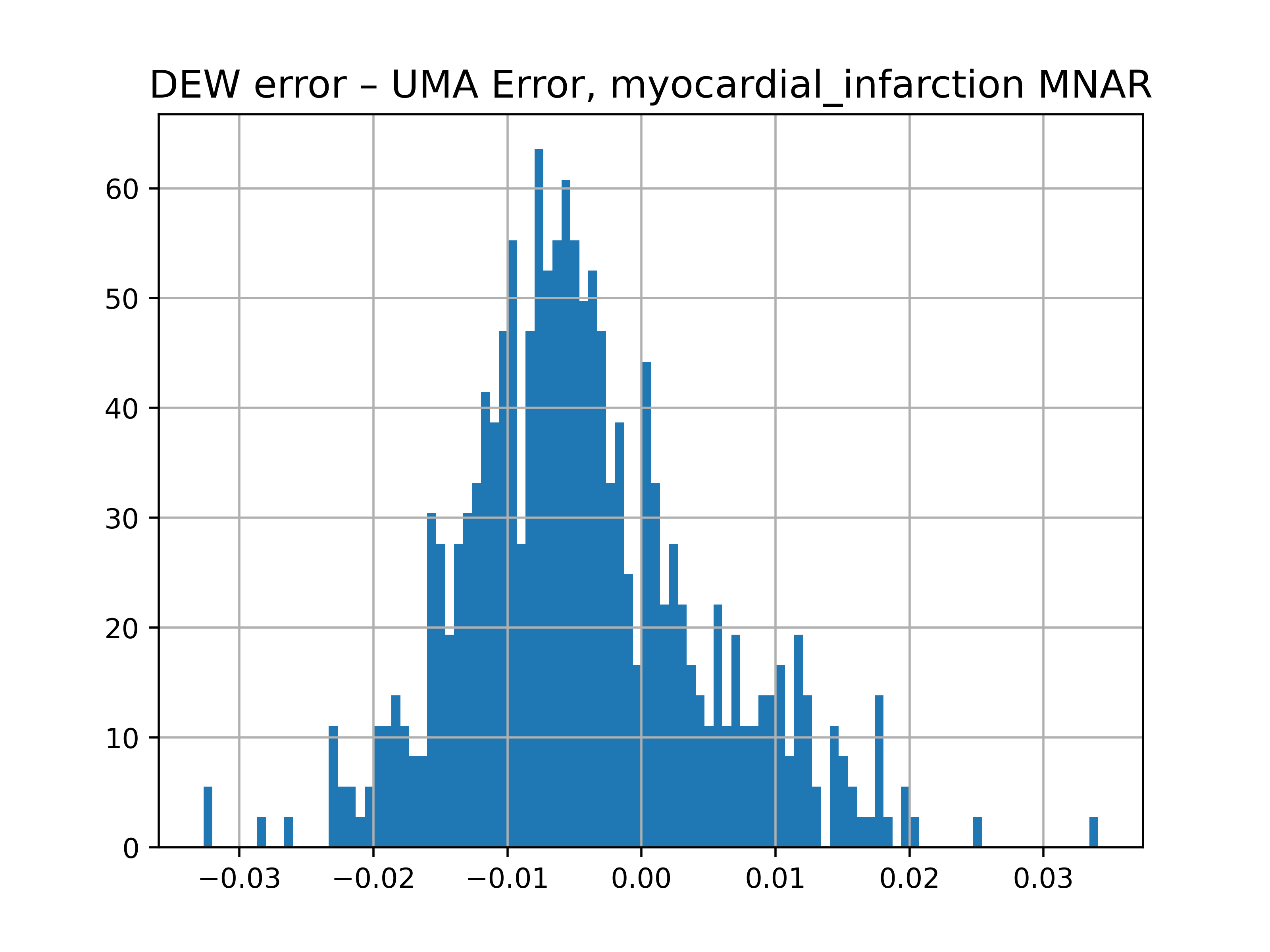}}
    }
\end{figure*}

\section{Conclusion and Future Work} 
\label{sec:conclusion}


We have designed and evaluated an approach to extend the classical notion of dynamic ensemble weighting to datasets with missing data. Whereas previous approaches would have required first imputing and then training classifiers on the imputed dataset, we develop a technique for directly optimizing at the level of joint imputation-prediction pipelines. This approach estimates a soft ordering of pipelines by calculating expected relative competences of each pipeline via the pipelines' performances on nearby samples from the training set. Using relative expected competence scores as weights to scale the pipelines' predictions reduces model perplexity compared with uniform soft voting over pipelines, evidenced by the paired t-test results in Table~\ref{tab:paired t-test}. It improves classification probability error in the majority of samples, leading to better calibrated prediction distributions. This can be useful when calculating disease risk scores or in scenarios where quantifying uncertainty is important.

The small overhead required by the algorithm allows it to be used as an AutoML module; instead of performing model selection, one can use all baseline models to form a stronger predictor. M-DEW also preserves any interpretability afforded by the baseline models; it provides a clear, nearest-neighbor approach to assigning competence-based weights, so for each sample one can trace the contributions of each pipeline via that sample's nearest neighbors' competences in M-DEW's second training stage. 

In extending the dynamic ensembling approach to handle missing values in the inputs, our work paves the way for future research along these lines. M-DEW makes no assumptions about underlying input structure, as long as there is a suitable distance measure between inputs and suitable objective function for outputs.  

We plan to extend the approach to utilize learned representations of the inputs. We are also investigating ways to jointly optimize the imputation and downstream prediction models using differentiable programs such as neural networks with combined loss functions.
\section{Acknowledgements}
Research reported in this publication was supported by the National Library Of Medicine of the National Institutes of Health under
Award Number R01LM013327. The content is solely the responsibility of the authors and does not
necessarily represent the official views of the National Institutes of Health.
\bibliography{reference}

\newpage
\appendix

\section{Technical Appendix}
\label{apd:technical appendix}
In the appendix we provide details for the baseline code (Section 1), hardware information(Section 2), and certainty analysis with reliability diagrams for M-DEW(Section 3). 

\subsection{Baseline Code}
Our baseline is generated with several \emph{sklearn} modules, \emph{Xgboost}, and \emph{R-miss-tastic}. 
\begin{itemize}
    \item \textbf{Uniform Model Averaging (UMA)}~This is a technique that gives equal weight to each individual model's prediction when making a final prediction. It serve as a baseline ensemble method. 

    \item \textbf{R-miss-tastic}~\citep{mayer2019r} This method generates missing values from non-missing datasets with customized distribution of missing data percentage. \url{https://github.com/R-miss-tastic/website/tree/master/static/how-to/python}
    \item \textbf{KNN}~\citep{troyanskaya2001missing} The method uses the weighted average of its K-nearest neighbors to impute the missing value. \url{https://scikit-learn.org/stable/modules/generated/sklearn.impute.KNNImputer.html}
    \item \textbf{Bayesian Ridge}~ \citep{MacKay1992bayesian,Tipping2001Sparse} This method is based on Bayesian linear regression and is designed to handle regression problems while taking into account uncertainty in the model parameters. \url{https://scikit-learn.org/stable/modules/generated/sklearn.linear_model.BayesianRidge.html}
    \item \textbf{Xgboost}~\citep{chen2016xgboost} Xgboost is applied in both imputation stage and prediction stage. It is the effective and efficient method to deal with missing data. \url{https://github.com/dmlc/xgboost/blob/36eb41c960483c8b52b44082663c99e6a0de440a/doc/index.rst}
    \item \textbf{Random Forest}~\citep{breiman2001random} This method is applied in both imputation stage and prediction stage. It is a meta estimator that employs averaging to increase predicted accuracy and reduce overfitting after fitting numerous decision tree classifiers to different dataset subsamples. \url{https://scikit-learn.org/stable/modules/generated/sklearn.ensemble.RandomForestRegressor.html#sklearn.ensemble.RandomForestRegressor} \url{https://scikit-learn.org/stable/modules/generated/sklearn.ensemble.RandomForestClassifier.html}
\end{itemize}

\subsection{Hardware Information}

\begin{itemize}
    \item \textbf{CPU}
        \begin{verbatim}
Model name:          Intel(R)Xeon(R)CPU 
                     E5-2687W v4@3.00GHz
CPU(s):              48
Thread(s) per core:  2
CPU max MHz:         3500.0000
L1d cache:           32K
L1i cache:           32K
L2 cache:            256K
L3 cache:            30720K
        \end{verbatim}
    \item \textbf{GPU}
        \begin{verbatim}
GPU0:               Quadro K620
GPU1:               NVIDIA GeForce 
    \end{verbatim}
\end{itemize}
\subsection{Relative Error Histograms for datasets}
\begin{figure*}[htbp]
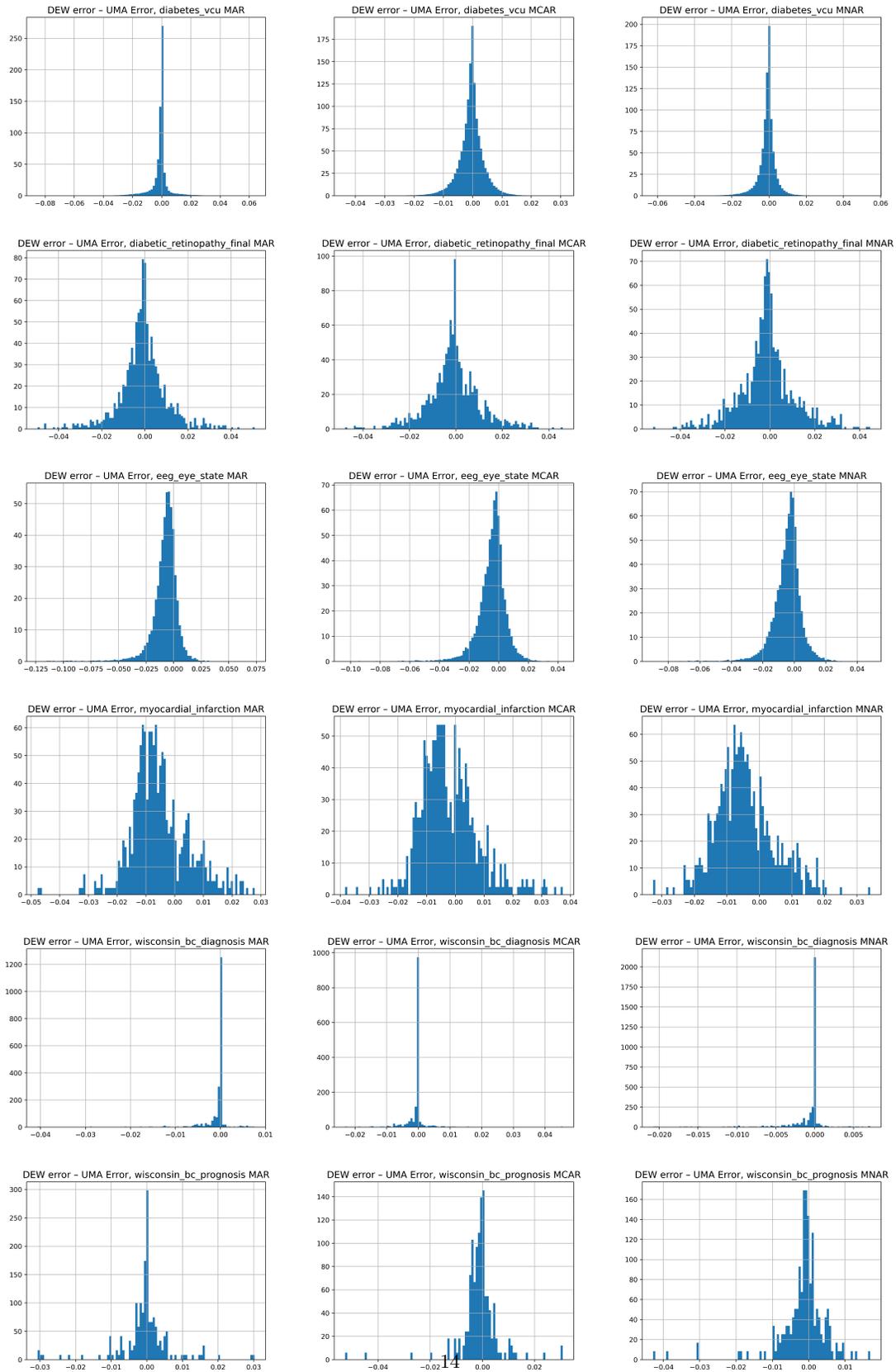

    \centering
    \caption{Relative Error Histograms} 
    \label{fig:rel_error_hist}

\foreach \x in \samplesB
  {\includegraphics[width=0.3\textwidth]{figures/relative_errors/\x}\hspace{0pt}{}}

\foreach \x in \samplesC
  {\includegraphics[width=0.3\textwidth]{figures/relative_errors/\x}\hspace{0pt}{}}

\foreach \x in \samplesD
  {\includegraphics[width=0.3\textwidth]{figures/relative_errors/\x}\hspace{0pt}{}}

\foreach \x in \samplesE
  {\includegraphics[width=0.3\textwidth]{figures/relative_errors/\x}\hspace{0pt}{}}

\foreach \x in \samplesF
  {\includegraphics[width=0.3\textwidth]{figures/relative_errors/\x}\hspace{0pt}{}}

\foreach \x in \samplesG
  {\includegraphics[width=0.3\textwidth]{figures/relative_errors/\x}\hspace{0pt}{}}

\end{figure*}

Plots of the histograms of relative error improvements can be seen in Figure~\ref{fig:rel_error_hist}. 
The plots show M-DEW Error $-$ UMA Error. Therefore, the more dense the plot is to the left of $0$, the better DEW performs relative to UMA.

\subsection{Calibration Analysis}
\emph{Calibration analysis}~\citep{platt2000probabilistic,niculescu-mizil2005predicting, zadrozny2002transforming} is normally applied after training to statistically measure the model's ability with prediction probability.  It is sophisticated when classifying a data point to a correct label is important. Motivated by M-DEW's ability to reduce perplexity, we extend it by adding a further layer of calibration on top.

CalibratedClassifierCV~\footnote{\url{https://scikit-learn.org/stable/modules/generated/sklearn.calibration.CalibratedClassifierCV.html}}~\citep{scikit-learn} is an effective tool for enhancing the performance of predictive algorithms, particularly in classification tasks. This algorithmic technique aims to improve the reliability of probabilistic predictions made by classifiers, ensuring they are well-calibrated and reflect the true likelihood of class membership. By combining a base classifier with a calibration model, CalibratedClassifierCV transforms the raw predicted probabilities into more accurate estimates, thus providing more meaningful confidence scores. 

 CalibratedClassifierCV can significantly enhance its predictive capabilities when applied to a new algorithm. The new algorithm can first generate raw predicted probabilities for each class, which are often optimistic and may not accurately represent the true class probabilities. Next, the calibrated model produces more accurate and well-calibrated probabilities, an S-shape curve. For the M-DEW algorithm, We experiment calibration analysis for all six datasets.

 There are three groups of reliability diagrams with Sigmoid calibration as Figure~\ref{fig:calibration_groups}. In plotting below, the x-axis represents the average predicted probability in each bin. The y-axis is the fraction of positives, i.e., the proportion of samples whose class is the positive class (in each bin). The term "bin" in refers to the number of "bins" or calibration intervals used to transform the raw predicted scores into calibrated probabilities. By dividing the predicted probabilities into bins, we can observe how well the predicted probabilities align with the actual fraction of positive samples within each bin of our model. The rest are the calibrated results(reliability diagrams).

The ideal shape of curves should be close to the S-shape and instead appear to be fluctuating for dataset Wisconsin Breast Cancer diagnosis and prognosis. It usually suggests that the models may have some degree of miscalibration or inconsistency in their predicted probabilities. However, those two datasets are not in S-shape because of sample size or data imbalance. Fluctuations in the calibration curve can also occur when the sample size is small, or there is a significant class imbalance in the data. The calibration curve may not smooth out in such cases due to limited data points in certain regions.

\begin{figure*}[!hpt]
    \centering
    \includegraphics[width=0.8\textwidth]{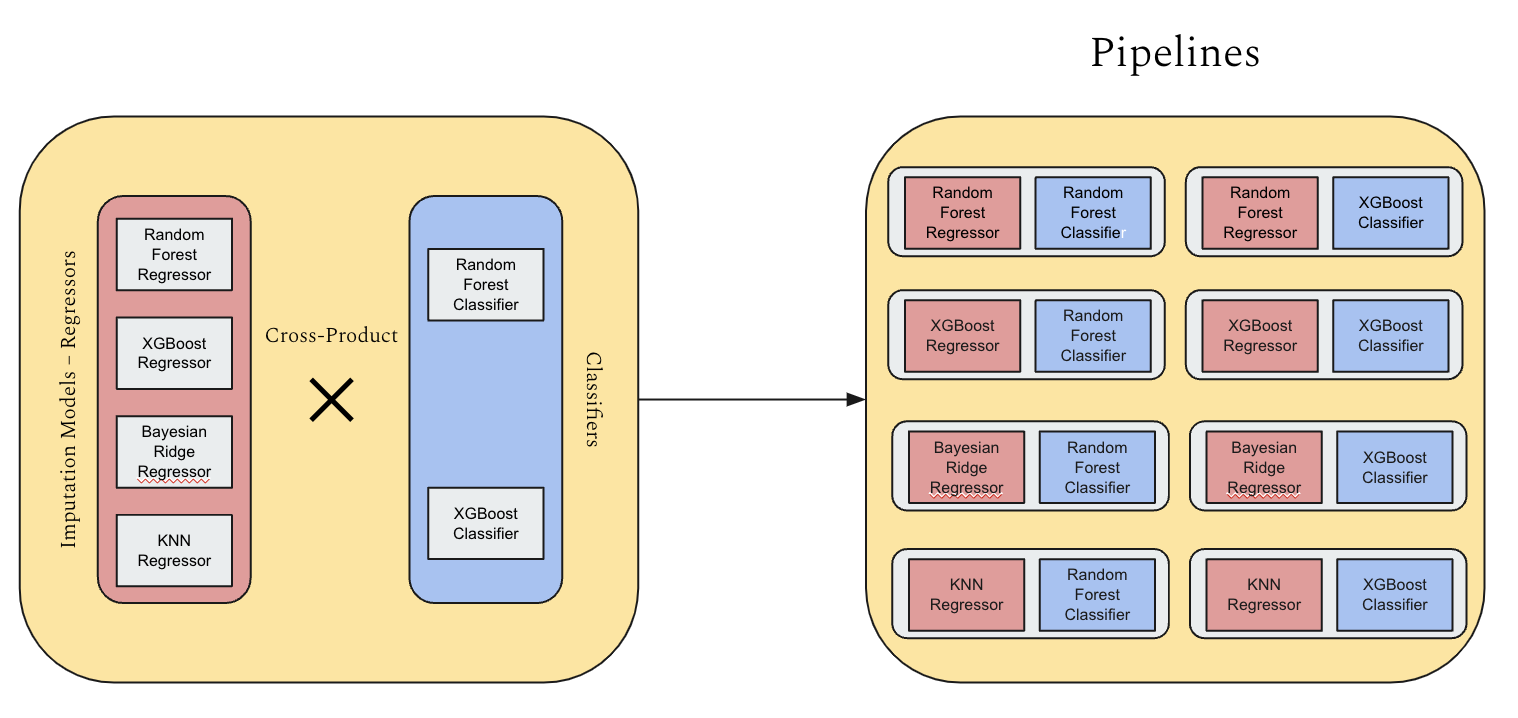}
    \caption{M-DEW's imputing-prediction pipeline. There are total 8 base estimators to construct M-DEW. Random Forest, XGBoosting, and KNN are popular imputation technique}
    \label{fig:regressors_classifiers}
\end{figure*}

\begin{table*}[!hpt]
    \centering 
    \sisetup{round-mode=places, round-precision=3}
    \resizebox{0.85\textwidth}{!}{
    \begin{tabular}{|l|l|l|}
        \hline
        Experiment & UMA & M-DEW \\ \hline
        eeg\_eye\_state MNAR & \num{0.25912537645573047} & 0.2601360786916317 \\ \hline
        eeg\_eye\_state MCAR & 0.2587541559455349 & 0.2596832891854269 \\ \hline
        eeg\_eye\_state MAR & 0.2599553776319339 & 0.26178878397810695 \\ \hline
        diabetic\_retinopathy MAR & 0.3020440141903601 & 0.3031740221628122 \\ \hline
        diabetic\_retinopathy MNAR & 0.2891785425641328 & 0.2902275287724473 \\ \hline
        diabetic\_retinopathy MCAR & 0.30269251203479697 & 0.30352194179421066 \\ \hline
        wisconsin\_bc\_prognosis MCAR & 0.19828763284593903 & 0.1984543459655042 \\ \hline
        wisconsin\_bc\_prognosis MNAR & 0.18263723211754626 & 0.18393224974616498 \\ \hline
        wisconsin\_bc\_prognosis MAR & 0.20945584159775288 & 0.2095635635424871 \\ \hline
        wisconsin\_bc\_diagnosis MCAR & 0.4237567782008261 & 0.42418764860161845 \\ \hline
        wisconsin\_bc\_diagnosis MNAR & 0.4238170852085251 & 0.4241030127687644 \\ \hline
        wisconsin\_bc\_diagnosis MAR & 0.4221709207468057 & 0.42236447517994835 \\ \hline
        myocardial\_infarction MCAR & 0.2954491814718572 & 0.2976526040103881 \\ \hline
        myocardial\_infarction MAR & 0.298894038746503 & 0.3027085660867358 \\ \hline
        myocardial\_infarction MNAR & 0.31277641034218473 & 0.3156143066211194 \\ \hline
        Diabetes\_vcu MAR & 0.2630907363015426 & 0.2637330086475949 \\ \hline
        Diabetes\_vcu MCAR & 0.25845379258736195 & 0.2587403688534198 \\ \hline
        Diabetes\_vcu MNAR & 0.26287367651477866 & 0.2634343516876172 \\ \hline
    \end{tabular}
    }
    \caption{Brier Score-a metric for evaluating the precision of probability predictions, particularly in binary classification tasks. During the development of M-DEW, the Brier Score offered trustworthy probability estimates, which is crucial for risk assessment and decision-making procedures. Dew's results outperformed UMA}
    \label{tab:Brier scores}
\end{table*}

\begin{table*}[!hpt]
  \centering 
  \sisetup{round-mode=places, round-precision=3}
  \resizebox{0.85\textwidth}{!}{
  \begin{tabular}{ll*{12}{>{\centering\arraybackslash}p{1cm}}}
  \toprule
   {\textbf{Dataset}} & \multicolumn{3}{c}{\textbf{UMA}} & \multicolumn{3}{c}{\textbf{M-DEW}}\\
    \cmidrule(lr){2-4} \cmidrule(lr){5-7} \cmidrule(lr){8-10} \cmidrule(lr){11-13} & \textbf{AUC} & \textbf{Acc.} & \textbf{F1} & \textbf{AUC} & \textbf{Acc.} & \textbf{F1} \\

    \midrule
    Eye-eeg-state MAR & \num{0.8939} & \num{0.8058}& \num{0.7639} & \num{0.8988 }& \num{0.8107 } & \num {0.7712}\\ 
    Eye-eeg-state MCAR & \num{0.8329 }& \num{0.7496 }&\num{ 0.6866 } & \num{0.8365 }& \num{0.7535 }&\num {0.6931}\\ 
    Eye-eeg-state MNAR & \num{0.8382 }& \num{0.7576 }&\num {0.6971 }& \num{0.8417 }&\num{0.7596 }& \num{0.7005}\\ 
    \hline
    Diabetic Retinopathy MAR& \num{0.7537 }& \num{0.6811 }& \num{0.6813}& \num{0.7311 }& \num{0.6637 }& \num{0.6639}\\ 
    Diabetic Retinopathy MCAR& \num{0.6658 }& \num{0.6333 }& \num{0.6334}& \num{0.6109 }& \num{0.5951 }& \num{0.5939}\\ 
    Diabetic Retinopathy MNAR& \num{0.7318 }& \num{0.6654 }& \num{0.6650 }& \num{0.7327 }& \num{0.6637 }& \num{0.6632}\\ 
    \hline
    Breast Cancer Wisconsin Diagnostic MAR& \num{0.9903} & \num{0.9473} & \num{0.9471} & \num{0.9904} & \num{0.9473 }& \num{0.9471}\\ 
    Breast Cancer Wisconsin Diagnostic MCAR& \num{0.9893 }& \num{0.9543 }& \num{0.9540} & \num{0.9893} & \num{0.9526 }& \num{0.9522}\\ 
    Breast Cancer Wisconsin Diagnostic MNAR& \num{0.9929 }& \num{0.9491 }& \num{0.9490}&  \num{0.9927} & \num{0.9473 }& \num{0.9472 }\\ 
    \hline
    Breast Cancer Wisconsin Prognostic MAR& \num{0.5439 }& \num{0.7420 }& \num{0.6899} & \num{0.5462 }& \num{0.7370 }& \num{0.6819}\\
    Breast Cancer Wisconsin Prognostic MCAR& \num{0.5886} & \num{0.7576 }& \num{0.6799} & \num{0.5894 }& \num{0.7576 }& \num{0.6799}\\
    Breast Cancer Wisconsin Prognostic MNAR& \num{0.6228} & \num{0.7626 }& \num{0.7094} & \num{0.6229} & \num{0.7626} & \num{0.7094}\\
    \hline
    
    Myocardial infarction MAR& \num{0.9024} & \num{0.8100} & \num{0.7950} & \num{0.9027} & \num{0.8137 }& \num{0.8006} \\
    Myocardial infarction MCAR& \num{0.8335} & \num{0.7397 }& \num{0.7279} & \num{0.8314 }& \num{0.7378 }& \num{0.7265}\\
    Myocardial infarction MNAR& \num{0.8916} & \num{0.7878} & \num{0.7790} & \num{0.8919} & \num{0.7878} & \num{0.7790}\\
    \hline
    Diabetes 130 hospitals MAR& \num{0.7124} & \num{0.6520} & \num{0.5724} & \num{0.7126} & \num{0.6526}& \num{0.5750}\\ 
    Diabetes 130 hospitals MCAR& \num{0.6773} & \num{0.6253} & \num{0.5129} & \num{0.6774} & \num{0.6254} & \num{0.5147}\\
    Diabetes 130 hospitals MNAR& \num{0.7082 }& \num{0.6492 }& \num{0.5640} & \num{0.7086} & \num{0.6495} & \num{0.5657}\\
    
    \bottomrule
  \end{tabular}
  }
    \caption{Standard classification metrics - AUROC (AUC), accuracy (Acc.), F1-score (F1),between UMA and M-DEW. We use AUROC to assess model discrimination between classes, accuracy to measure the overall correctness of predictions, and the F1 score to balance precision and recall, particularly in imbalanced datasets. In most cases, M-DEW performed better than UMA. 
}
    \label{tab:metrics} 
\end{table*}

\end{document}